\documentclass[smallcondensed]{svjour3}     
\usepackage{graphicx}
\usepackage{xcolor}
\usepackage{float}
\usepackage{amssymb}
\usepackage{microtype}
\usepackage{subfigure}
\usepackage{booktabs} 
\usepackage{hyperref}
\usepackage{amsmath}
\usepackage{mathtools}

\usepackage{algorithm}
\usepackage{algpseudocode}

\newcommand{\greencheck}{{\color{green}\checkmark}}
\newcommand{\orangecheck}{{\color{orange}\checkmark}}
\newcommand{\redx}{{\color{red}x}}

\begin{document}


\title{From MNIST to ImageNet and Back: Benchmarking Continual Curriculum Learning}

\titlerunning{M2I and I2M: Benchmarking Continual Curriculum Learning}










\author{Kamil Faber \and
        Dominik Zurek \and
        Marcin Pietron \and
        Nathalie Japkowicz \and
        Antonio Vergari$^{*}$ \and
        Roberto Corizzo$^{*}$
}


\institute{Kamil Faber, Dominik Zurek, Marcin Pietron \at
              AGH University of Science and Technology, Cracow, 30059, Poland \\
              \{kfaber,dzurek,pietron\}@agh.edu.pl
           \and
           Antonio Vergari \at
              University of Edinburgh, Edinburgh, EH8 9AB, UK \\
              avergari@ed.ac.uk
          \and
           Nathalie Japkowicz, Roberto Corizzo \at
           American University, Washington, DC, 20016, USA\\
           \{japkowic,rcorizzo\}@american.edu\\
           \\
           * Equal Supervision \\
           Corresponding Author: Roberto Corizzo \email{rcorizzo@american.edu} 
}

\date{Received: date / Accepted: date}

\maketitle
\abstract{Continual learning (CL) is one of the most promising trends in recent machine learning research. Its goal is to go beyond classical assumptions in machine learning and develop models and learning strategies that present high robustness in dynamic environments. This goal is realized by designing strategies that simultaneously foster the incorporation of new knowledge while avoiding forgetting past knowledge. The landscape of CL research is fragmented into several learning evaluation protocols, comprising different learning tasks, datasets, and evaluation metrics. Additionally, the benchmarks adopted so far are still distant from the complexity of real-world scenarios, and are usually tailored to highlight capabilities specific to certain strategies. In such a landscape, it is hard to clearly and objectively assess models and strategies. In this work, we fill this gap for CL on image data by introducing two novel CL benchmarks that involve multiple heterogeneous tasks from six image datasets, with varying levels of complexity and quality. Our aim is to fairly evaluate current state-of-the-art CL strategies on a common ground that is closer to complex real-world scenarios. We additionally structure our benchmarks so that  tasks are presented in increasing and decreasing order of complexity -- according to a curriculum -- in order to evaluate if current CL models are able to exploit structure across tasks. We devote particular emphasis to providing the CL community with a rigorous and reproducible  evaluation protocol for measuring the ability of a  model to generalize and not to forget while learning.
Furthermore, we provide an extensive experimental evaluation showing that popular CL strategies, when challenged with our proposed benchmarks, yield sub-par performance, high levels of forgetting, and present a limited ability to effectively leverage curriculum task ordering. 
We believe that these results highlight the need for rigorous comparisons in future CL works as well as pave the way to design new CL strategies that are able to deal with more complex scenarios.
}




\section{Introduction}
\label{sec:intro}

Continual Learning (CL), also known as Lifelong Learning, is a promising learning paradigm to design models that have to learn how to perform \textit{multiple tasks} across different environments over their lifetime \cite{parisi2019continual} \footnote{To uniform the language and enhance the readability of the paper we adopt the unique term continual learning (CL).}.
%
Ideal CL models in the real world should be able to quickly adapt to new environments and tasks, while perfectly retaining what they learned in the past, thus only increasing, and not decreasing, their performance as they experience more tasks.
In practice, this is quite challenging due to the hardness of generalizing from one environment to another when there is a huge \textit{distribution shift} between them \cite{lopezpaz2017,krawczyk2021tensor,li2017learning,cano2022rose}, and to the fact that models tend to (sometimes catastrophically) \textit{forget} what they learned for previous tasks.

The great attention around this paradigm has brought many communities to focus on how to address these challenges, including reinforcement learning \cite{baker2023domain} \cite{abel2018policy} and anomaly detection \cite{faber2022active} \cite{corizzo2022cpdga}.
However, the majority of attention has been devoted to computer vision, generating a Cambrian explosion of CL models \cite{lopezpaz2017,li2017learning,aljundi2018memory,kang2022,chaudhry2019,zenke2017,rolnick2019experience,hihn2022hierarchically}, where the most common task is to learn models that can classify different kinds of images while preventing catastrophic forgetting or quickly adapting to new image classes or image datasets.
Every new model has been evaluated in a slightly different setting -- \textit{using a different dataset, evaluation metrics and learning protocols} -- thus generating a number of CL learning and evaluation schemes. 
%
The result is that \textit{the benchmark panorama of CL in computer vision is quite fragmented}, and therefore it has become tougher to measure catastrophic forgetting and domain adaptation in a fair and homogeneous way for the many CL models we have in the literature these days. 
Furthermore, all previous evaluation protocols are designed to highlight some specific model characteristics and, as such, are generally over-simplified w.r.t. real-world data \cite{cossu2022class}.

For example, one of the most popular evaluation protocol for CL models in computer vision is to design different tasks to classify different (subsets of the) classes of a single dataset \cite{DeLange2022ACL,van2019three}. 
The most prominent example is splitMNIST in which the 10 digits from MNIST \cite{lecun1998mnist} are (usually) divided into 5 tasks consisting of 2 digits each. 
Similar approaches are proposed for CIFAR10 \cite{Krizhevsky2009LearningML}, and TinyImagenet \cite{le2015tiny}.
Other datasets, such as Continuous Object Recognition (CORe50) \cite{pmlr-v78-lomonaco17a}, specifically designed for LL, still make the same assumptions to generate tasks.
Clearly, these protocols are not suited to detect distribution shifts, due to the high inter-task similarity. 
Consequently, catastrophic forgetting is much easier to prevent in these cases. 
Therefore the reported metrics for models evaluated in this way can be overly optimistic.


To deal with domain shifts, researchers have recently started to sample tasks from two different datasets. 
For instance, \cite{lopezpaz2017} proposed to train and evaluate a model on Imagenet first and then challenge its performance on the Places365 dataset. 
\cite{li2017learning} considers more scenarios, starting with Imagenet or Places365, and then moving on to the VOC/CUB/Scenes datasets. 
%
Few works propose more advanced scenarios built on top of more than two datasets. The two most prominent examples are the so-called 5-datasets \cite{ebrahimi2020adversarial} and RecogSeq \cite{aljundi2018}, which provide models with more challenging scenarios than previous attempts, increasing the number of considered datasets to 5 and 8, respectively. Unfortunately, those datasets provide a similar task complexity due to the limited differences across datasets.
Furthermore, when different datasets are employed, it is important to ``calibrate the meaning'' of the employed metrics, taking into account the number of classes involved in each task.
Despite all this progress, we argue that there is still not a robust and standardized evaluation benchmark for the many CL models in the literature.
We argue that a modern benchmark for CL should provide the following aspects. 
First, \textbf{\textit{multiple heterogeneous tasks}} that do not restrict to a single set of concepts, e.g., digits in MNIST or SVHN  or naturalistic images as in Imagenet or CIFAR10.
Second, \textbf{\textit{a varying quality and complexity of the tasks}}, e.g. alternating from black and white (B\&W) to RGB images and vice-versa, considering different image sizes, and a number of concepts.
Third, a way to systematically evaluate if \textbf{\textit{learning on a curriculum}} of task complexities help with domain generalization and catastrophic forgetting.
For example, evaluating if a model trained on B\&W digits can better generalize to B\&W letters and then to RGB digits and letters, or if learning them in the inverse order is more beneficial.
Fourth, \textbf{\textit{a rigorous way to measure generalization and forgetting}} in terms of modern backward and forward transfer metrics \cite{diaz2018don} in a number of different evaluation scenarios, i.e., when classes or tasks are introduced incrementally \cite{van2019three}.  
Lastly, all results should be \textit{\textbf{exactly reproducible out-of-the-box}}. 
We argue that all the previous CL works discussed above do not consider one or more of these criteria, as highlighted in Table \ref{tab:benchmarks_comparison_2}. 
In addition to the five desiderata, we also cover both class and task-incremental learning settings, which is not usually the case for other surveyed works.
In this paper, we aim to overcome these limitations.

\begin{figure*}[h]
\centering
\includegraphics[width=1.0\linewidth]{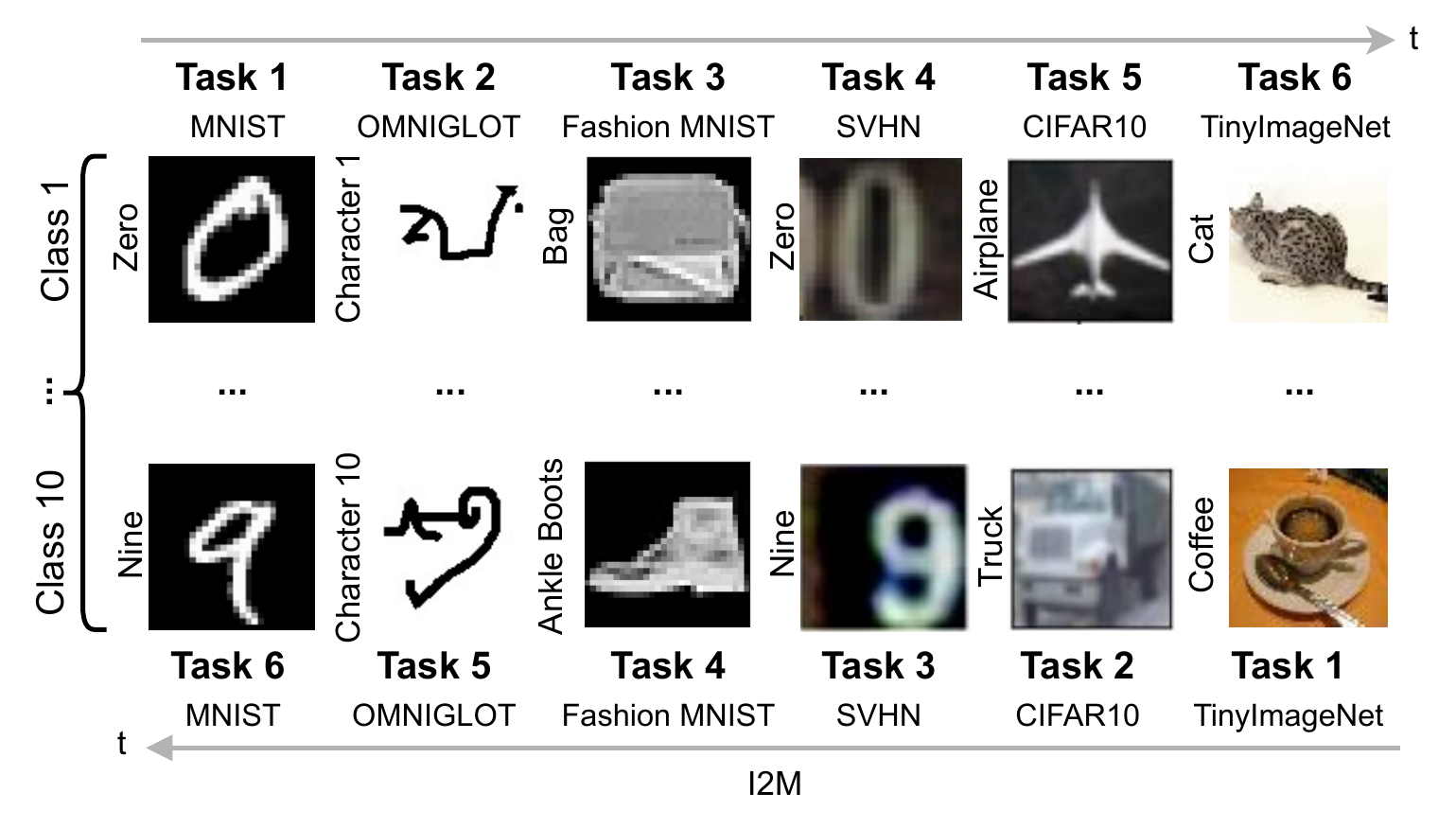}
\caption{The Proposed M2I and I2M continual learning benchmarks. There are 6 different tasks, each sampled from a different dataset: MNIST \cite{lecun1998mnist}, OMNIGLOT \cite{lake2015human}, Fashion MNIST \cite{xiao2017fashion}, SVHN \cite{Netzer2011ReadingDI}, CIFAR10 \cite{Krizhevsky2009LearningML} and TinyImageNet \cite{le2015tiny}. Tasks are organized in two curriculum ordering, from simple to harder (left to right) and backward (right to left). Every task sports 10 classes, as to make the performance metric meaning intuitive and faithful.}
\label{fig:workflow}
\end{figure*}

\begin{table}[t]
    \centering
      \caption{Benchmarks comparison considering only multi-dataset benchmarks. Columns refer to: i) supporting multiple heterogeneous tasks; ii) varying task complexity and quality; iii) evaluating curriculum strategies; iv) rigorous way to measure generalization and forgetting, and v) exactly reproducible out-of-the-box. In addition, we consider the coverage of class (CI) and task-incremental (TI) learning settings. 
    The symbols have the following meaning: \greencheck - criterion is covered; \orangecheck - criterion is covered at some part or with some limitations; \redx - criterion is not covered at all.}
    \begin{tabular}{lccccccc}
    \toprule
    Benchmark        &  i) & ii) & iii) & iv) & v) & CI) & TI) \\
    \midrule
    Imagenet/Places365 to  VOC/CUB/Scenes \cite{li2017learning}     & \orangecheck & \redx &  \redx  & \orangecheck & \orangecheck & \redx & \greencheck  \\
    Imagenet to Places365  \cite{mallya2017} & \orangecheck & \redx & \redx & \orangecheck & \greencheck  & \redx & \greencheck \\
    5-Datasets        \cite{ebrahimi2020adversarial}  & \greencheck & \orangecheck  & \redx & \greencheck &  \greencheck  & \greencheck & \redx \\ 
    RecogSeq \cite{aljundi2018}      \cite{DeLange2022ACL} & \greencheck & \orangecheck & \redx & \orangecheck & \greencheck & \greencheck & \greencheck \\
    \midrule
    M2I, I2M \textit{(ours)}              & \greencheck  & \greencheck  & \greencheck  & \greencheck & \greencheck & \greencheck & \greencheck \\
    \bottomrule
    \end{tabular}
  
    \label{tab:benchmarks_comparison_2}
\end{table}
Specifically, the contributions of the paper are as follows:
\begin{itemize}
    \item We 
    propose \textit{a set of benchmarks} built on 6 image datasets ordered in a curriculum of complexity -- \textit{from MNIST to TinyImageNet} (M2I) and back \textit{from TinyImageNet to MNIST} (I2M) -- that simultaneously satisfies all the above desiderata. 
    These benchmarks have varying task complexity, starting with simple digits and going to complex naturalistic images and viceversa (see Figure \ref{fig:workflow});
    \item  We provide an exhaustive experimental evaluation including 9 state-of-the-art continual learning methods, covering the key categories of approach (architectural, regularization, and rehearsal) in both class and task-incremental settings, and evaluating results using the most recent metrics adopted in the continual and lifelong learning community. 
\end{itemize}

\section{Background}
\label{sec:background}

\subsection{CL scenario types}
\label{sec:scenario_types}
%
A wide range of scenarios was designed and discussed in recent studies to design effective CL models while trying to reflect real-world challenges. 
In image classification, two main scenarios are the most widely adopted: i) \textbf{\textit{task-incremental}} \cite{de2021continual} and ii) \textbf{\textit{class-incremental}} \cite{belouadah2021comprehensive} LL. 
%
In both scenarios, the model has to learn new tasks, which are presented sequentially.
Each incoming task provides the model with new, previously unseen classes, that need to be incorporated.

A common characteristic for both mentioned scenarios is the availability of task boundaries, which make the CL method aware that a new task is presented. 
The most relevant difference between the two scenarios is the availability of task labels, which provide the model with additional information on which task is being processed at the moment, during both training and inference. 
Specifically, a task-incremental scenario assumes the availability of task labels, whereas in class incremental learning, this information is not available.

\textbf{\textit{It is worth stressing that the same data presented in different types of scenarios can yield significantly different results}}, since certain CL methods may be tailored for task-incremental scenarios, and as such the exploitation of task labels improve their performance, while they may significantly suffer in class-incremental scenarios, where this information is not available.

The most widely adopted benchmark for class and task-incremental scenarios is split-MNIST \cite{kirk2017} consisting of 5 tasks. It leverages the original MNIST separating it into five tasks, each containing two digits. In this class-incremental scenario, the model is not aware of whether what the current task is. It is only aware of the fact that it encountered a new task and needs to adjust itself. During the testing phase, the model is also not informed about which set of digits is currently provided, so the model has to classify one of the ten classes (digits 0-9). On the other hand, in the task-incremental variant of split-MNIST, the model is aware of which task is currently being presented, and only decides whether the image belongs to the first or the second class of the current task. This prediction, combined with information about the current task id, leads to the specific digit prediction.

Less commonly, certain scenarios relax the assumptions of class and task-incremental scenarios \cite{lomonaco2019nicv2}. 
Notable examples include domain-incremental scenarios \cite{baker2023domain} where new distributions of the same classes are presented over time, as well as task-agnostic scenarios, where neither task labels nor task boundaries are not available, and reliance on external methods is necessary to detect task changes \cite{faber2022lifewatch}. 

\subsection{CL Strategies}
\label{sec:sota}

From a broad perspective, CL strategies belong to three main groups: using \textit{\textbf{regularization}}, \textit{\textbf{dynamic architectures}}, and \textit{\textbf{rehearsal}} (also known as experience replay). 
In this paper, we consider popular and largely adopted CL strategies. We now describe each strategy and the rationale for its adoption in the benchmarks.

Regularization strategies influence the model weights adjustment process that takes place during model training in the attempt to preserve knowledge of previously learned tasks. 
The regularization strategies considered include Elastic Weight Consolidation (EWC) \cite{kirk2017}, Learning without Forgetting (LwF), Synaptic Intelligence (SI) \cite{zenke2017}, and Memory Aware Synapses (MAS) \cite{aljundi2018memory}.
LwF \cite{li2017learning} aims at achieving  output stability through knowledge distillation. When a new task is observed, the new model is incentivized to predict values that are close to the outputs of the model learned prior to this task.
EWC \cite{kirk2017} and SI \cite{zenke2017} adopt a weighted quadratic regularization loss which penalizes moving weights that are important for previous tasks. The EWC loss is based on the Fisher Information Matrix which presents a higher computational complexity than the surrogate loss used in the SI method. 
Similarly, Memory Aware Synapses (MAS) \cite{aljundi2018memory} estimates the cumulative importance of model weights as new tasks are encountered, penalizing changes to weights that are crucial for previously learned tasks.  
Shifting the focus on dynamic architectures, CWRStar \cite{lomonaco2019nicv2} adapts weights exclusively for the last layer before the prediction layer, freezing all previous layers. 
Finally, rehearsal strategies considered include GDumb \cite{prabhu2020}, Replay \cite{rolnick2019experience}, Gradient Episodic Memory (GEM) \cite{lopezpaz2017}, and Average Gradient Episodic Memory (AGEM) \cite{chaudhry2019}.
GDumb \cite{prabhu2020} is a greedy strategy that stores samples for all classes in a buffer, and uses them to iteratively retrain a model from scratch.
Replay \cite{rolnick2019experience} follows a similar approach, but stores a balanced number of samples per task, which are used to fine-tune previously trained models.
A more sophisticated approach is pseudo-rehearsal with generative models. GEM \cite{lopezpaz2017} is a fixed-size memory that stores a subset of old patterns and influences the loss function through inequality constraints. 
AGEM \cite{chaudhry2019} is a revised version of GEM that performs averaging to increase efficiency.

The rationale for the adoption of the  aforementioned strategies in our benchmark is that they are heterogeneous in terms of approaches, and are particularly prevalent in the CL community. They represent the foundations in the CL field, and are often used to assess the competitiveness of emerging CL methods with respect to consolidated and diversified approaches. Moreover, they are easy to use and favor reproducibility, thanks to publicly available tools such as the Avalanche library \cite{lomonaco2021avalanche}.

\subsection{CL evaluation protocol and metrics}
\label{sec:metric_and_evaluation_protocol}
The standard evaluation procedure applied in continual image classification assumes the availability of a set of tasks, each defined with a set of classes.  The learning scenario consists of $N$ tasks $T = t_1, t_2, \dots, t_n$ where the model has to learn new tasks without forgetting previous tasks.



    


Metrics in continual learning usually focus on assessing the performance of a model (e.g., its accuracy) with respect to (at least one of) three crucial properties:
\textit{i)} performance on newly encountered tasks; \textit{ii)} performance retention capabilities on previously learned tasks (i.e., the \textit{ability to avoid or mitigate forgetting}); and \textit{iii)} knowledge transfer from learned tasks to new ones (i.e., \textit{the ability to generalize over newly occurring challenges}).
The first works in CL  proposed three metrics: average accuracy, backward transfer, and forward transfer to measure the above desiderata~\cite{lopezpaz2017}. However, in their original definition, only the performance of the model after learning all tasks was considered.

Instead, we consider model performances for all tasks and at all stages of the learning process, as understanding how performance changes before and after every task can provide several insights into the strength and weaknesses of every model \cite{diaz2018don}.
%
%
For simplicity, we will be storing the 
partial model performance, measured as classification accuracy, in a matrix $\mathsf{R}$ whose entries $\mathsf{R}_{i, j}$ represent the accuracy on a given task $j$ after learning task $i$. 

%
%

\textbf{Average Accuracy ($\mathsf{ACC}$)} -- It measures the average accuracy of the model after learning each task, evaluating only  the current and all previously learned tasks: 
    \begin{equation}
        \mathsf{ACC} = \sum\nolimits_{i \ge j}^N \mathsf{R}_{{i,j}}/ (N(N-1)/2),
    \end{equation}
defined as the average performance over all tasks the model has seen so far.

\vspace{3pt}

\textbf{Backward Transfer ($\mathsf{BWT}$)} -- It measures the impact of learning new tasks on the performance of all previously learned tasks. Negative backward transfer indicates that learning a new task is harmful to the performance of previously learned tasks (this issue is known as forgetting):
    \begin{equation}
        \mathsf{BWT} = \sum\nolimits_{i=2}^N\sum\nolimits_{j=1}^{i-1} (\mathsf{R}_{i, j} - \mathsf{R}_{j,j})/ (N(N-1)/2),
        \label{eq:bwt_all}
    \end{equation}
defined as the average amount of forgetting presented by the model on the overall scenario.

\vspace{3pt}

\textbf{Forward Transfer ($\mathsf{FWT}$)} -- It measures the impact of learned tasks on the performance of tasks learned in the future: 
    \begin{equation}
        \mathsf{FWT} = \sum\nolimits_{i<j}^{N} \mathsf{R}_{i, j}/(N(N-1)/2),
        \label{eq:fwt_all}
    \end{equation}
defined as the average model performance on yet unseen tasks.

\vspace{2pt}

\section{Our benchmarks: M2I and I2M}






\label{sec:benchmark}



In Section~\ref{sec:intro} we pointed out essential desiderata for continual learning benchmarks that are designed to reflect real-life environments and challenges. 
In the following, we further elaborate on each criterion, providing a rationale for its importance, and we describe how our benchmark tackles these challenges. 

First, it is important to \textbf{\textit{consider multiple heterogeneous tasks}}. The rationale is that, since continual learning models should adapt to new and unprecedented situations, as human beings usually act in real environments, they should be evaluated on sequences of heterogeneous tasks. While common benchmarks focus on homogeneous  tasks, such as different classes of handwritten digits (e.g. as in splitMNIST), heterogeneous tasks have the advantage of reflecting more realistic cases where the model is challenged by unprecedented tasks with great diversity.
To deal with multiple heterogeneous tasks,
our benchmark leverages 6 largely-varying image classification datasets: MNIST (handwritten digits) \cite{lecun1998mnist}, Omniglot (alphabets) \cite{lake2015human}, Fashion MNIST (clothing items) \cite{xiao2017fashion}, SVHN (street view house numbers) \cite{Netzer2011ReadingDI}, CIFAR10 (small real-world images) \cite{Krizhevsky2009LearningML}, and TinyImagenet (multi-domain large-scale real-world images) \cite{le2015tiny}. Each dataset is regarded as a task, resulting in a learning scenario with six tasks  with heterogeneous characteristics. 
We provide more details about the datasets included and the preprocessing they underwent into the benchmark in Table \ref{tab:datasets}.

Second, it is important to devise scenarios with \textbf{\textit{varying quality and task complexity}}, since an ideal model should present generalization capabilities dealing with easy, moderate, and difficult tasks at the same time, as found in the real-world. 
Ideal scenarios should avoid simplistic sequences of tasks with high task similarity, and prefer introducing new tasks that are different enough from the previous one, thus challenging the model in a significant way. 
%
This aspect should comprise having tasks on images varying in terms of visual and chromatic quality and difficulty of classification.
Our benchmarks take this into consideration as they include very complex multi-domain real-world image classification such as TinyImagenet (harder classification), as well as handwritten digit recognition in MNIST (easier classification), and letter recognition in different alphabets in Omniglot (moderate difficulty).
This choice of datasets creates ambitious but realistic challenges for CL strategies, allowing us to test their limitations.

Third, the hardness of each task is relative to the ordering in which the task is presented to the model.
E.g., task ordering is important for us humans as we do not learn challenging new tasks from scratch but, instead, incrementally build up the necessary skills to perform these new tasks, leveraging a combination of skills learned in the past.
We would require the same efficiency from a continual learner.
%
Therefore it is crucial to evaluate models \textit{\textbf{learning on a direct or inverse curriculum}}.
 The adoption of direct curriculum learning -- learning on tasks of increasing complexity -- in conventional machine learning research showcased that significant improvements in generalization can be achieved, increasing the speed of convergence of the training process \cite{bengio2009curriculum} \cite{gao2022learning} \cite{song2020ada}. 

When it comes to LL, however, direct and inverse curriculum learning are overlooked.
Indeed, in the best cases, multiple random task orderings are provided in addition to a single task order. 
To properly consider curriculum learning, our benchmark considers curriculum learning by devising a task order according to their difficulty. 
The scenario starts with MNIST (black \& white handwritten digits), which is regarded as an easy task. The following tasks are Omniglot (alphabets) and Fashion MNIST (clothing items), which present a spike of complexity compared to MNIST. Subsequently, SVHN (street view house numbers) brings real-world complexity by introducing images gathered from cameras with colors. CIFAR10 presents the same challenges of real-world colored images and extends them with more challenging patterns encountered in complex objects. Finally, the highest level of complexity is provided by multi-domain large-scale images from TinyImagenet.
In addition to the direct curriculum direction where tasks are ordered as described (from MNIST to TinyImageNet, aka M2I), we also cover the opposite case of decreasing order of difficulty (from TinyImageNet to MNIST, aka I2M). 


Fourth, \textbf{\textit{rigorous way to measure generalization and forgetting}}.
The most important aspect of continual machine learning is to design strategies and models that are able to incorporate new tasks during their lifespan, without forgetting previous tasks. 
Metrics such as $\mathsf{BWT}$ and $\mathsf{FWT}$ are introduced for this reason, see Section~\ref{sec:metric_and_evaluation_protocol}
However, they can be cumbersome to interpret or lose their meaning, depending on the learning setting at hand.
%
%
For instance, $\mathsf{FWT}$ is ill-defined in a class-incremental scenario since the model will never predict classes that were never presented before.
Another example is that of multi-dataset benchmarks where tasks contain a varying number of classes.
Specifically, tasks with a reduced number of classes will exhibit a random performance that is higher (e.g., 0.5 for 2 classes) than tasks with a higher number of classes (e.g., 0.1 for 10 classes).
As results are generally aggregated (i.e., averaged) across tasks \cite{li2017learning} \cite{mallya2017}, \cite{ebrahimi2020adversarial}, \cite{aljundi2018}, \cite{DeLange2022ACL}, CL metrics will be hard to interpret due to a different reference point for random performance. Ideal benchmarks should take these aspects into consideration to make sure that the calculation and the interpretation of results are correct.

To consider this aspect, we designed each task in our benchmark to contain 10 classes. In the case of MNIST and Fashion MNIST, SVHN, and CIFAR10, we use all classes. In the case of TinyImageNet and Omniglot, we select 10 classes. For TinyImagenet, we use Egyptian cat; reel; volleyball; rocking chair; lemon; bullfrog; basketball; cliff; espresso; plunger.
As for Omniglot, we select classes corresponding to  characters from the Alphabet of the Magi.
This setting allows us to preserve a high interpretability of all the resulting metric values
overcoming the limitation of tasks with imbalanced number of classes, where interpretability can be lost. 
Furthermore, to deal with class imbalance, we align the size of majority classes to that of minority classes. By doing so, we isolate the learning setting and avoid typical issues that arise in imbalanced learning, which might undermine the analysis of the final results.

Fifth, \textbf{\textit{exactly reproducible out-of-the-box}}. Many benchmarks are not reproducible due to the lack of precise details on model configurations and experimental settings. This issue is exacerbated when the code is unavailable and it is required to implement the scenario and the evaluation scheme from scratch. In other cases, when the code is available, it is not general enough to be leveraged in different settings, e.g. when comparing with the latest models and strategies. 
To this end, our benchmark is implemented on top of Avalanche \cite{lomonaco2021avalanche} -- the state-of-the-art open-source library for LL. This choice ensures the reproducibility of the experiments and paves the way for the adoption and extension of the benchmark for future research. The code for our benchmarks is publicly available at the following repository URL: \url{https://github.com/lifelonglab/M2I_I2M_benchmark}.

\begin{table}[]
    \caption{Overview of original datasets involved in our benchmarks. The datasets present heterogeneous characteristics, i.e., domains and technical quality.
    For TinyImagenet, we select the following classes: Egyptian cat; reel; volleyball; rocking chair; lemon; bullfrog; basketball; cliff; espresso; plunger.
    As for Omniglot, we select characters from the alphabet of the Magi.
    }
    \label{tab:datasets}
    \centering
    \begin{tabular}{lllrr}
    \toprule
    Dataset     & Colors    & Size      & Classes    & Available images \\
    \midrule
    MNIST           & BW        & 28x28     & 10        & 70 000 \\
    Omniglot        & BW        & 105x105   & 1632      & 32 460  \\
    Fashion MNIST   & BW        & 28x28     & 10        & 70 000  \\
    SVHN            & RGB       & 32x32    & 10        & 630 420  \\
    CIFAR10         & RGB       & 32x32     & 10        & 60 000  \\
    TinyImagenet    & RGB       & 64x64     & 200       & 100 000  \\
    \bottomrule
    \end{tabular}
\end{table}

\section{Experiments and discussion}
\label{sec:exp}
We carry out experiments involving both the \textit{task-incremental} and \textit{class-incremental} CL scenario types described in Section \ref{sec:scenario_types}, the CL strategies devised in Section \ref{sec:sota}, and the CL evaluation protocol and metrics defined in Section \ref{sec:metric_and_evaluation_protocol}. 
%
We run an exhaustive series of experiments on our proposed M2I and I2M benchmarks for LL, resulting in 88 complete experiments (considering M2I and I2M with 11 CL strategies, 2 learning settings, and 2 model backbones) and 528 runs (model training and evaluation).
We aim to answer the following research questions: 
\begin{itemize}
    \item \textbf{RQ1)} Do our benchmarks provide challenging scenarios for state-of-the-art CL strategies as discussed in Section~\ref{sec:sota}? That is, are these strategies still as accurate and robust w.r.t the metrics defined in Section~\ref{sec:metric_and_evaluation_protocol} as originally introduced in their papers when exposed to M2I and I2M?
    \item \textbf{RQ2)}
    Can state-of-the-art CL strategies leverage direct and indirect curriculum task ordering to maximize their backward and forward transfer? Or alternatively, do different task orderings with varying task complexity have an impact on the final performance?
\end{itemize}

%
We first detail the experimental setup of our experiments and then provide an in-depth discussion of the results gathered. 
For the curious reader, the short answer to both questions is that overall the state-of-the-art models underperform when executed on our challenging benchmarks, despite many of these models were supposed to be robust to catastrophic forgetting and multiple tasks. 

\subsection{Experimental setup}

As mentioned in Section \ref{sec:benchmark}, our benchmark provides multiple heterogeneous tasks with varying quality and task complexity. For instance, 3 of the 6 tasks contain black and white images, whereas the remainder contain colored images. Moreover, the image size varies across all tasks. There may be different ways to deal with different image channel types and sizes, which can have an impact on the final performance. However, we recognize that finding the optimal solution is an open challenge for researchers working with our benchmark, and it is out of the scope of this paper. For simplicity, for image sizes, we adopt the most frequently adopted approach, which consists in resizing all images to the same size ($64 \times 64$). To deal with different image channels, we consider the largest number of channels (RGB) for all tasks ($3$) and replicate the single-channel encountered in BW images to all $3$ channels.
We recall that, in order to provide a rigorous way to measure generalization and forgetting, we balance class sizes by taking $500$ images from each of them, for both the training and evaluation phases. By doing so, we isolate possible issues deriving from class imbalance from our evaluation. 

\paragraph{Network architecture.}
We leverage two commonly used model backbones in CL with different parameter sizes, as to measure the effect of overparametrization w.r.t. our performance metrics in LL.
Each network architecture is being used across all strategies.
We employ a Wide VGG9 \cite{simonyan2014very} as a smaller neural network for image data and an EfficientNet-b1 \cite{tan2019efficientnet} as a larger alternative. 
The hyperparameter configuration used in the experiments is: \textit{\{ epochs=50, learning\_rate=0.001, momentum=0.9 \}}. Optimization takes place through Stochastic Gradient Descent (SGD) using the Cross-Entropy loss. 
We experimented with different negative powers of $10$ for the configuration of the learning rate as suggested in \cite{bengio2012practical}, 
For the number of epochs, we experimented with similar values to those reported in the original publications of CL strategies \cite{rolnick2019experience}\cite{aljundi2018}. 
Preliminary experiments showed that different configurations did not provide a significant difference in terms of performance metric values.

\paragraph{CL strategies.}
In addition to the state-of-the-art CL strategies covered by our experiments and described in Section \ref{sec:sota}, we adopt two additional baseline approaches which loosely correspond to lower and upper bound model performance:
\begin{itemize}
\item \textbf{Naive (fine-tuning)}: The model is incrementally fine-tuned without considering any mechanism to preserve past knowledge, which, in principle, should yield a high degree of forgetting. 
This strategy allows us to compare the performance (in terms of accuracy) and forgetting (in terms of backward transfer) of smarter CL strategies.
\item \textbf{Cumulative}: New data is accumulated as it comes, and the model is retrained using all available data. 
The rationale for this baseline is to simulate upper-bound performance assuming full knowledge of the data, and unlimited computational resources to deal with stored data (storage) and model retraining (time). Cumulative can also be regarded as a variant of Replay with unlimited memory. This baseline is interesting since it allows us to estimate the accuracy that could be achieved at a much higher computational cost.
\end{itemize}
For technical details on the hyperparameter setting of the CL strategies are provided in our above-mentioned GitHub repository, which includes the code to reproduce our experiments.

\subsection{Discussion: RQ1}

%
%
We present our results both as aggregated metrics computed after all the tasks have been learned in Tables \ref{tab:results_vgg9_mnist_img} -- \ref{tab:efficient_net_not_pretrained_I2M} as well as disaggregated accuracy results evaluating every task past task after learning a new one (as entries in the matrix $\mathsf{R}$, see Section~\ref{sec:metric_and_evaluation_protocol}) as heatmaps shown in Figures \ref{fig:heatmaps_vgg9_mnist_img_class} -- \ref{fig:heatmaps_efficientnet_img_mnist_task}.
These heatmaps allow us to understand at a finer grain what are the failure modes of a strategy and whether certain tasks are harder than another. 
We now discuss four different settings, comprising either a class-incremental or task-incremental scenario and two task orderings (M2I or I2M). We start from employing the smaller model backbone: VGG9.

\subsubsection{VGG9}

For both M2I and I2M (see Tables \ref{tab:results_vgg9_mnist_img} -- \ref{tab:results_vgg9_img_mnist}), aggregated $\mathsf{ACC}$, $\mathsf{BWT}$ and $\mathsf{FWT}$ (only for task-incremental scenarios, see Section~\ref{sec:benchmark}) are disappointing for all strategies discussed in Section~\ref{sec:sota}.
They do not achieve a positive backward transfer and all highlight a systematic catastrophic forgetting, while the forward transfer floats around chance level (10\%\footnote{We remark that this trend is easy to spot and understand in our benchmarks as all tasks sports only 10 classes.}).
Staple strategies such as LwF, MAS, GDUMB and SI
 are generally comparable with the Naive strategy (i.e., just applying fine-tuning). 
%
%

We inspect rankings over the $\mathsf{ACC}$ score to see if any notable trend manifests between different scenario types.
We found that  
AGEM is very weak in the class-incremental setting (ranked 11 and 9, respectively), but quite robust in the task-incremental setting (ranked 4 and 3). Surprisingly, GEM seems to be robust in both settings (ranked 3 in three out of four settings, and 8 in the fourth setting). CWRStar achieves a moderately high position in the class-incremental setting (ranked 4 in the ranking for M2I). 
While its ranking is surprisingly high, we remark that the raw performance is clearly unsatisfactory, when considered in absolute terms in this context. 
A much lower ranking is observed in the task-incremental setting, despite the slight improvement in its performance. Overall, CWRStar appears ineffective in preventing catastrophic forgetting across all tasks in our settings, and it appears that it just focuses on memorizing the first task.

The method that seems to be the least prone to forgetting is Replay. This result is surprising since the total memory size chosen for the replay buffer in the experiment is just 200 samples. 
It is also interesting to observe that Replay presents a performance that is quite close to Cumulative (an $8\%-18\%$ decrease in accuracy across the four mentioned settings) using a fraction of available data (less than 1\%). 
As expected, Cumulative presents the best performance across all four learning settings. However, \textit{it should be seen as an unrealistic upper bound}, since it assumes that infinite memory and training time are allowed for the model. On a different note, the positive results confirm that the scenarios designed in our benchmark are reasonable and can be, in principle, learned by the model but current CL strategies.
While these have shown to be reliable in conventional CL scenarios, they are not bulletproof and present limited robustness when exposed to more complex scenarios, such as our M2I (\textbf{RQ1}).

\begin{figure*}[h]
        \centering
        \includegraphics[width=\textwidth]{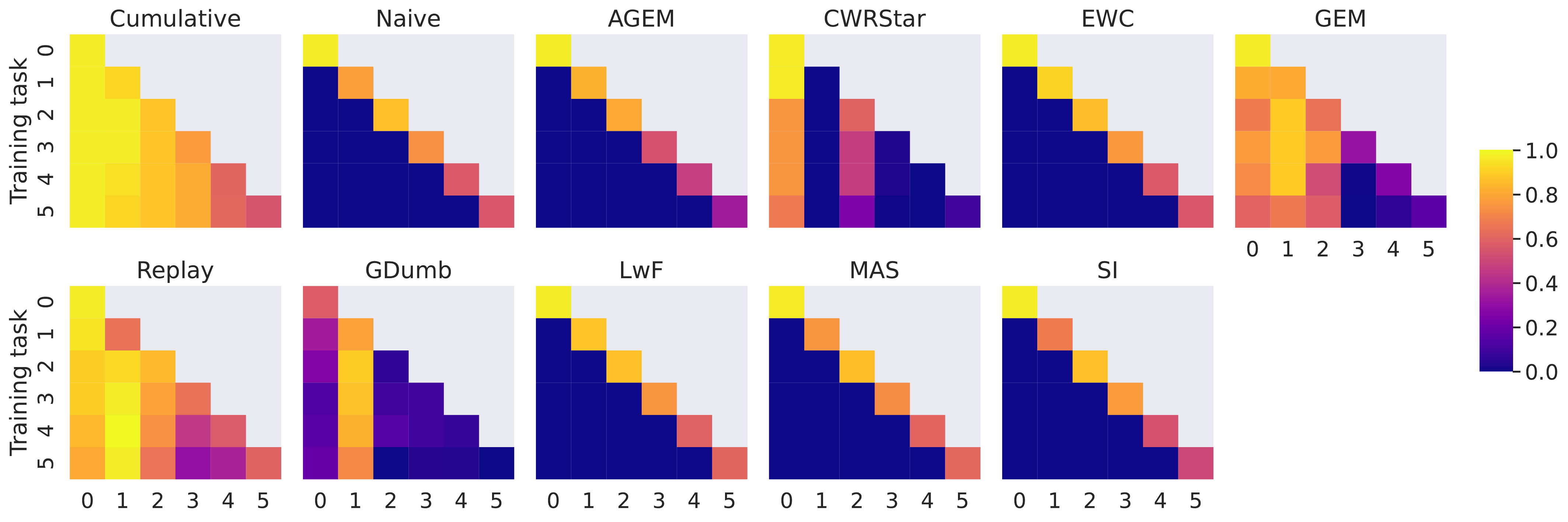}
        \caption{Experimental results (Wide-VGG9 -- M2I -- Class-incremental) in terms of disaggregated performance ($\mathsf{ACC}$) on single tasks after learning previous tasks.
        }
        \label{fig:heatmaps_vgg9_mnist_img_class}
    \end{figure*}

\begin{figure*}[h]
        \centering
        \includegraphics[width=\textwidth]{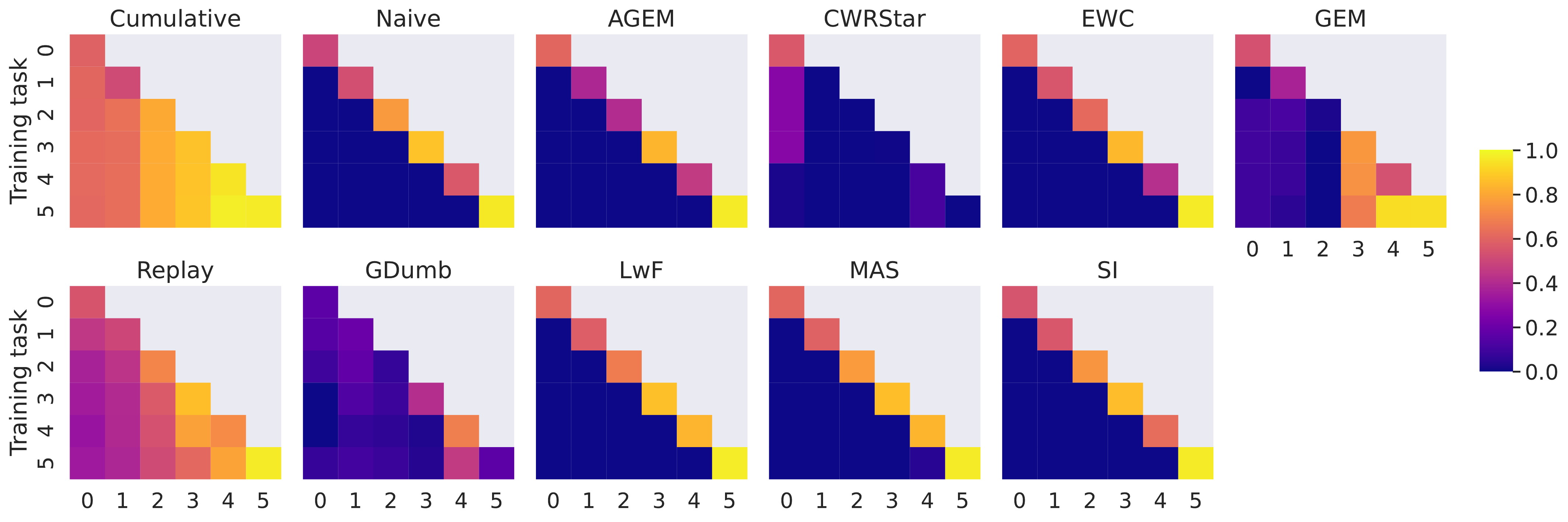}
        \caption{Experimental results (Wide-VGG9 -- I2M -- Class-incremental) in terms of disaggregated performance ($\mathsf{ACC}$) on single tasks after learning previous tasks. 
        }
        \label{fig:heatmaps_vgg9_img_mnist_class}
    \end{figure*}

\begin{figure*}[h]
        \centering
        \includegraphics[width=\textwidth]{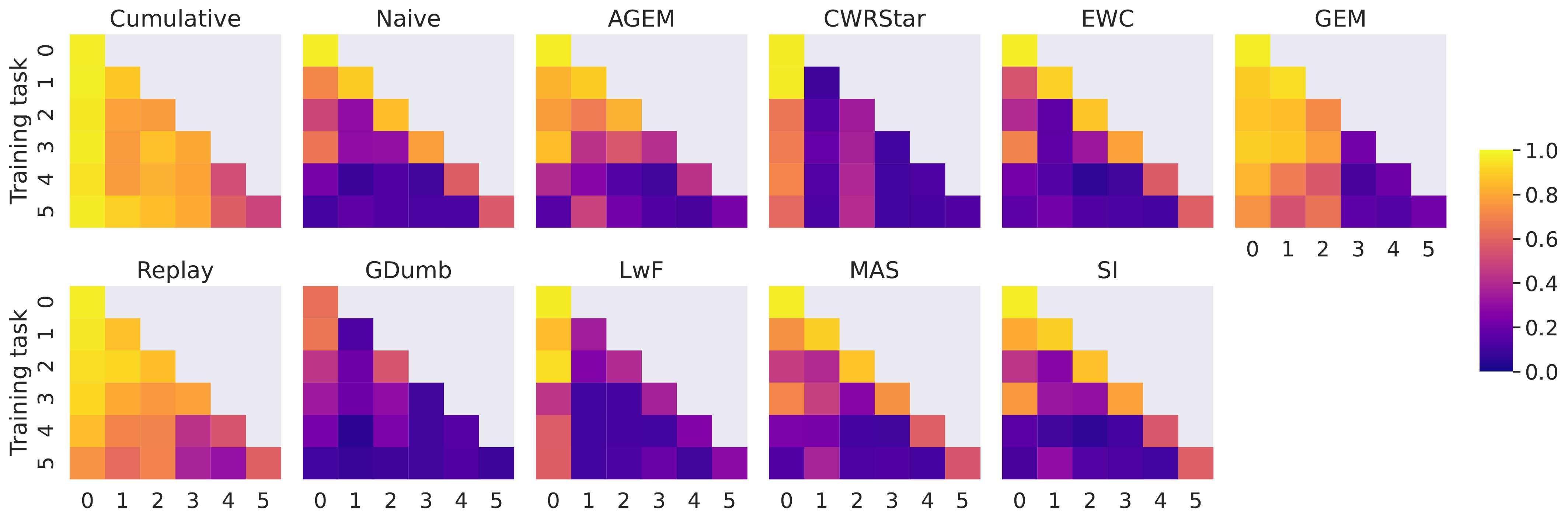}
        \caption{Experimental results (Wide-VGG9 -- M2I -- Task-incremental) in terms of disaggregated performance ($\mathsf{ACC}$) on single tasks after learning previous tasks.}
        \label{fig:heatmaps_vgg9_mnist_img_task}
    \end{figure*}

\begin{figure*}[h]
        \centering
        \includegraphics[width=\textwidth]{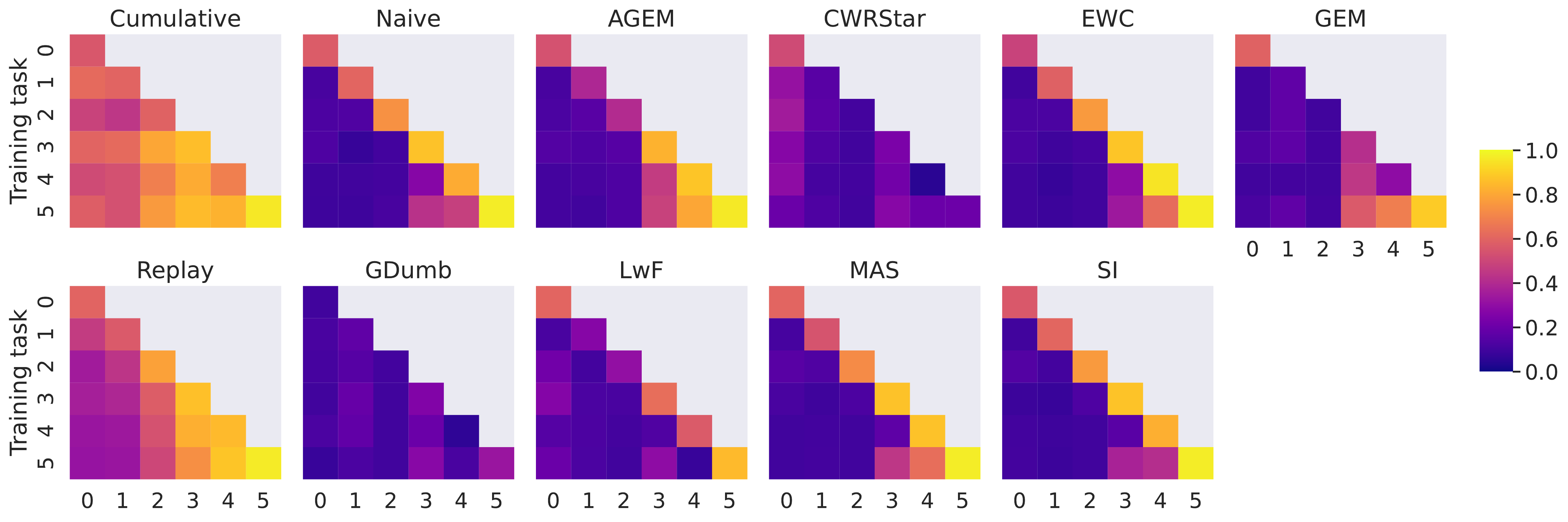}
        \caption{Experimental results (Wide-VGG9 -- I2M -- Task-incremental) in terms of disaggregated performance ($\mathsf{ACC}$) on single tasks after learning previous tasks.}
        \label{fig:heatmaps_vgg9_img_mnist_task}
    \end{figure*}

Observing the heatmaps in Figure \ref{fig:heatmaps_vgg9_mnist_img_class} -- \ref{fig:heatmaps_vgg9_img_mnist_task}  allows us to zoom in and pinpoint the performance drops of different strategies on specific tasks. In this context, observing a decreasing performance on previously learned tasks is a clear manifestation of forgetting.
Results are quite negative for M2I in the class-incremental setting (see Figure \ref{fig:heatmaps_vgg9_mnist_img_class}). GEM preserves a good performance until the third task is presented, and then dramatically drops in the following tasks, due to their increasing complexity. GDumb presents a high performance on the second task throughout the entire scenario, but an unsatisfactory performance on all other tasks. This behavior likely depends on the fact that it is possible to learn the second task (Omniglot) with a limited number of samples, whereas this is too difficult for all other tasks. All other strategies, except Replay and Cumulative, struggle to preserve the knowledge of previous tasks and are successful at learning the last task exclusively, as evidenced by the very low-performance scores. 
The results for M2I in the task-incremental setting (see Figure \ref{fig:heatmaps_vgg9_mnist_img_task}) showcase a higher overall performance, with lower forgetting than the class-incremental setting. For instance, it can be observed that MAS and SI is able to preserve much more knowledge for some tasks, while its forgetting was rather drastic in the class-incremental setting. 

For I2M in the class-incremental setting (see Figure \ref{fig:heatmaps_vgg9_img_mnist_class}), a similar behavior to the class-incremental counterpart of M2I can be observed, with drastic forgetting, which is even worse than the M2I scenario.  
As for I2M in the task-incremental setting (see Figure \ref{fig:heatmaps_vgg9_img_mnist_task}), it is also interesting to observe worse results than in the M2I task-incremental setting. 
This is also counterintuitive, as successfully learning a harder task should provide the model with enough knowledge not to perform so poorly on much simpler tasks that, as MNIST, might require learning only simple edge detectors.
We conjecture that this behavior might depend on the fact that once a model is presented with very different but complex tasks earlier in the scenario (e.g., ImageNet and CIFAR10) it might have a harder time learning to abstract useful features for simpler tasks later.
%


    \begin{table}[]
        \centering
        \begin{tabular}{lrrrrr}
\toprule
 & \multicolumn{2}{c}{Class-incremental} & \multicolumn{3}{c}{Task-incremental} \\
\cmidrule(lr){2-3} \cmidrule(lr){4-6} 
 & $\mathsf{ACC}$ & $\mathsf{BWT}$ & $\mathsf{ACC}$ & $\mathsf{BWT}$ & $\mathsf{FWT}$\\
\midrule
AGEM & 0.188 (11) & -0.241 & 0.472 (4) & -0.161 & 0.101 \\
Cumulative & 0.868 (1) & 0.004 & 0.819 (1) & 0.012 & 0.102 \\
CWRStar & 0.324 (4) & -0.044 & 0.356 (9) & -0.019 & 0.094 \\
EWC & 0.220 (7) & -0.271 & 0.395 (8) & -0.224 & 0.102 \\
GEM & 0.572 (3) & -0.074 & 0.613 (3) & -0.053 & 0.099 \\
GDumb & 0.304 (5) & -0.040 & 0.235 (11) & -0.071 & 0.092 \\
LwF & 0.222 (6) & -0.270 & 0.349 (10) & -0.083 & 0.096 \\
MAS & 0.215 (8) & -0.260 & 0.440 (5) & -0.212 & 0.100 \\
Naive & 0.213 (9) & -0.261 & 0.411 (7) & -0.228 & 0.093 \\
Replay & 0.755 (2) & -0.038 & 0.730 (2) & -0.086 & 0.101 \\
SI & 0.206 (10) & -0.255 & 0.419 (6) & -0.219 & 0.107 \\
\bottomrule
\end{tabular}

        \caption{Experimental results (Wide-VGG99 -- M2I) in terms of average performance (and rank) for all CL strategies in two learning settings.}
        \label{tab:results_vgg9_mnist_img}
    \end{table}
    
    \begin{table}[]
        \centering
        \begin{tabular}{lrrrrr}
\toprule
 & \multicolumn{2}{c}{Class-incremental} & \multicolumn{3}{c}{Task-incremental} \\
\cmidrule(lr){2-3} \cmidrule(lr){4-6} 
 & $\mathsf{ACC}$ & $\mathsf{BWT}$ & $\mathsf{ACC}$ & $\mathsf{BWT}$ & $\mathsf{FWT}$\\
\midrule
AGEM & 0.175 (9) & -0.180 & 0.345 (3) & -0.094 & 0.131 \\
Cumulative & 0.735 (1) & 0.012 & 0.663 (1) & 0.018 & 0.120 \\
CWRStar & 0.079 (11) & -0.037 & 0.202 (10) & -0.011 & 0.107 \\
EWC & 0.190 (8) & -0.202 & 0.340 (5) & -0.161 & 0.126 \\
GEM & 0.297 (3) & -0.031 & 0.269 (8) & 0.005 & 0.117 \\
GDumb & 0.155 (10) & -0.052 & 0.147 (11) & 0.000 & 0.085 \\
LwF & 0.216 (5) & -0.238 & 0.262 (9) & -0.105 & 0.088 \\
MAS & 0.223 (4) & -0.241 & 0.342 (4) & -0.148 & 0.130 \\
Naive & 0.199 (7) & -0.215 & 0.334 (6) & -0.160 & 0.131 \\
Replay & 0.550 (2) & -0.047 & 0.571 (2) & -0.061 & 0.135 \\
SI & 0.205 (6) & -0.223 & 0.323 (7) & -0.170 & 0.134 \\
\bottomrule
\end{tabular}

        \caption{Experimental results (Wide-VGG99 -- I2M) in terms of average performance (and rank) for all CL strategies in two learning settings.}
        \label{tab:results_vgg9_img_mnist}
    \end{table}
    
\subsubsection{EfficientNet}
Results on M2I and I2M (see Tables \ref{tab:efficient_net_not_pretrained_M2I} -- \ref{tab:efficient_net_not_pretrained_I2M}) show that the Naive strategy (ranked 5) achieves a performance that is close to some of the CL strategies (e.g. MAS, SI, EWC) but is significantly inferior to top performing strategies (Replay, Cumulative). 
When comparing class-incremental and task-incremental settings for M2I, some methods appear significantly more robust in the latter, with a simultaneous increase in their performance and position in the ranking\footnote{It is important to track both aspects, since the task-incremental setting is fundamentally easier than class-incremental, and observing only the absolute performance of the methods is not indicative of an improvement.}. This is the case for AGEM (ranked 9 and 5, respectively),  and GEM (ranked 4 and 3, respectively). For I2M, comparing class and task-incremental settings, the same phenomenon can be observed for a larger number of methods AGEM (ranked 9 and 6, respectively), MAS (ranked 8 and 7, respectively), SI (ranked 7 and 4, respectively), and EWC (ranked 6 and 5, respectively). 
Some strategies present a rather stable behavior in the two learning settings, since they appear to preserve their ranking. For M2I, this is the case for GDumb, LwF, and Replay. For I2M, this phenomenon applies to CWRStar, GDumb, and Replay.
As observed in VGG9 results, Cumulative showcases the best performance across all four learning settings, resulting in the top-ranked strategy.
Therefore, in absolute terms, the performance of informed strategies can be regarded as unsatisfactory, as it appears significantly lower than Cumulative. 
This result suggests that even increasing the parametrization of the backbone model does not provide these staple CL strategies to significantly improve over our simpler baselines in complex benchmarks such as our M2I  and I2M (\textbf{RQ1}).

    \begin{table}[h]
        \centering
        \begin{tabular}{lrrrrr}
\toprule
 & \multicolumn{2}{c}{Class-incremental} & \multicolumn{3}{c}{Task-incremental} \\
\cmidrule(lr){2-3} \cmidrule(lr){4-6} 
 & $\mathsf{ACC}$ & $\mathsf{BWT}$ & $\mathsf{ACC}$ & $\mathsf{BWT}$ & $\mathsf{FWT}$\\
\midrule
AGEM & 0.179 (9) & -0.224 & 0.328 (5) & -0.171 & 0.091 \\
Cumulative & 0.834 (1) & 0.003 & 0.656 (1) & 0.025 & 0.101 \\
CWRStar & 0.326 (3) & -0.014 & 0.367 (4) & 0.001 & 0.092 \\
EWC & 0.180 (8) & -0.222 & 0.272 (9) & -0.199 & 0.106 \\
GEM & 0.312 (4) & -0.056 & 0.420 (3) & -0.043 & 0.096 \\
GDumb & 0.029 (11) & -0.007 & 0.099 (11) & 0.000 & 0.100 \\
LwF & 0.165 (10) & -0.205 & 0.240 (10) & -0.124 & 0.096 \\
MAS & 0.187 (6) & -0.190 & 0.276 (8) & -0.198 & 0.104 \\
Naive & 0.205 (5) & -0.257 & 0.290 (6) & -0.220 & 0.103 \\
Replay & 0.655 (2) & -0.041 & 0.605 (2) & -0.098 & 0.102 \\
SI & 0.184 (7) & -0.231 & 0.277 (7) & -0.205 & 0.099 \\
\bottomrule
\end{tabular}

        \caption{Experimental results (EfficientNet -- M2I) in terms of average performance (and rank) for all CL strategies in two learning settings.}
        \label{tab:efficient_net_not_pretrained_M2I}
    \end{table}

    \begin{table}[h]
        \centering
        \begin{tabular}{lrrrrr}
\toprule
 & \multicolumn{2}{c}{Class-incremental} & \multicolumn{3}{c}{Task-incremental} \\
\cmidrule(lr){2-3} \cmidrule(lr){4-6} 
 & $\mathsf{ACC}$ & $\mathsf{BWT}$ & $\mathsf{ACC}$ & $\mathsf{BWT}$ & $\mathsf{FWT}$\\
\midrule
AGEM & 0.157 (9) & -0.134 & 0.289 (6) & -0.101 & 0.122 \\
Cumulative & 0.606 (1) & 0.021 & 0.596 (1) & 0.015 & 0.118 \\
CWRStar & 0.053 (10) & -0.035 & 0.171 (10) & -0.020 & 0.103 \\
EWC & 0.197 (6) & -0.182 & 0.299 (5) & -0.169 & 0.100 \\
GEM & 0.218 (3) & -0.052 & 0.276 (8) & -0.052 & 0.119 \\
GDumb & 0.029 (11) & 0.000 & 0.100 (11) & 0.000 & 0.092 \\
LwF & 0.203 (4) & -0.182 & 0.241 (9) & -0.072 & 0.091 \\
MAS & 0.178 (8) & -0.164 & 0.284 (7) & -0.148 & 0.105 \\
Naive & 0.202 (5) & -0.193 & 0.305 (3) & -0.154 & 0.108 \\
Replay & 0.417 (2) & -0.019 & 0.438 (2) & -0.058 & 0.099 \\
SI & 0.193 (7) & -0.177 & 0.302 (4) & -0.143 & 0.105 \\
\bottomrule
\end{tabular}

        \caption{Experimental results (EfficientNet -- I2M) in terms of average performance (and rank) for all CL strategies in two learning settings.}
        \label{tab:efficient_net_not_pretrained_I2M}
    \end{table}

\begin{figure}[h]
        \centering
        \includegraphics[width=\textwidth]{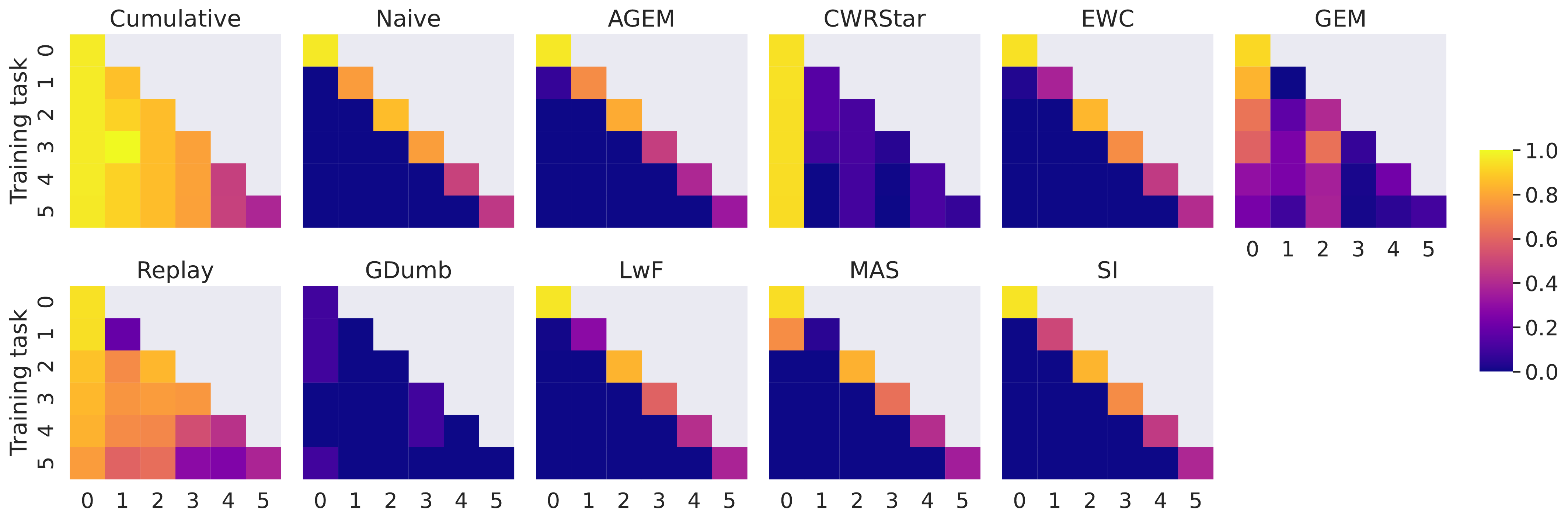}
        \caption{Experimental results (EfficientNet -- M2I -- Class-incremental) in terms of disaggregated performance ($\mathsf{ACC}$) on single tasks after learning previous tasks.}
        \label{fig:heatmaps_efficientnet_mnist_img_class}
    \end{figure}

\begin{figure}[h]
        \centering
        \includegraphics[width=\textwidth]{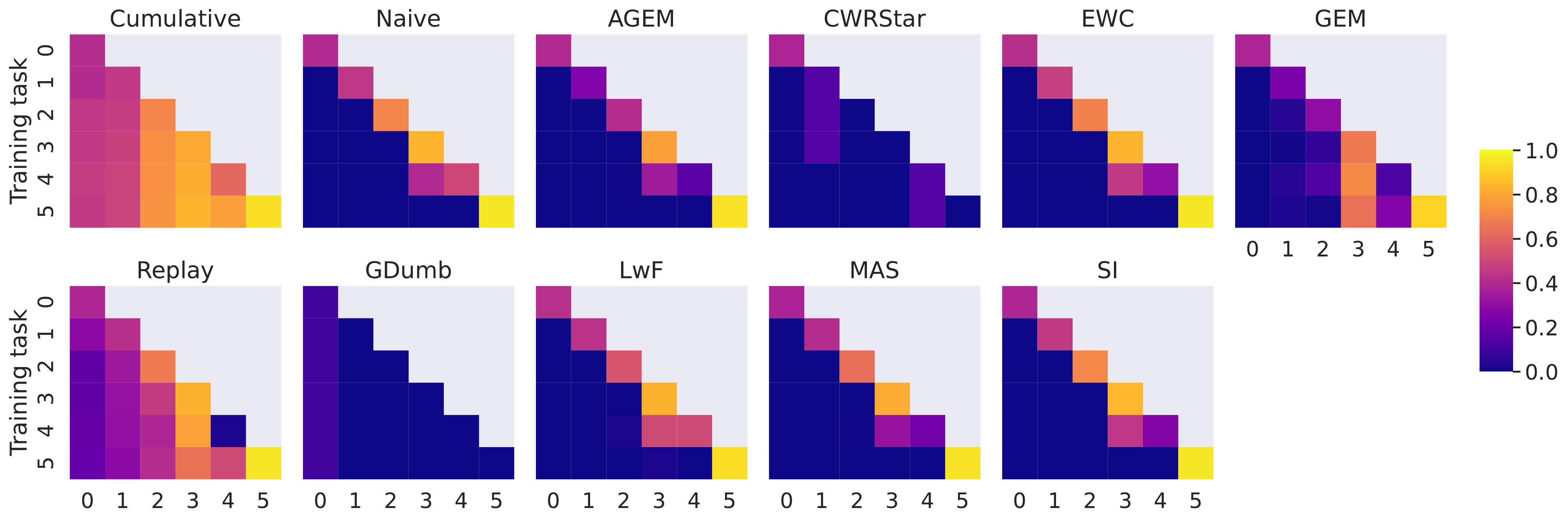}
        \caption{Experimental results (EfficientNet -- I2M -- Class-incremental) in terms of disaggregated performance ($\mathsf{ACC}$) on single tasks after learning previous tasks.}
        \label{fig:heatmaps_efficientnet_img_mnist_class}
    \end{figure}

\begin{figure}[h]
        \centering
        \includegraphics[width=\textwidth]{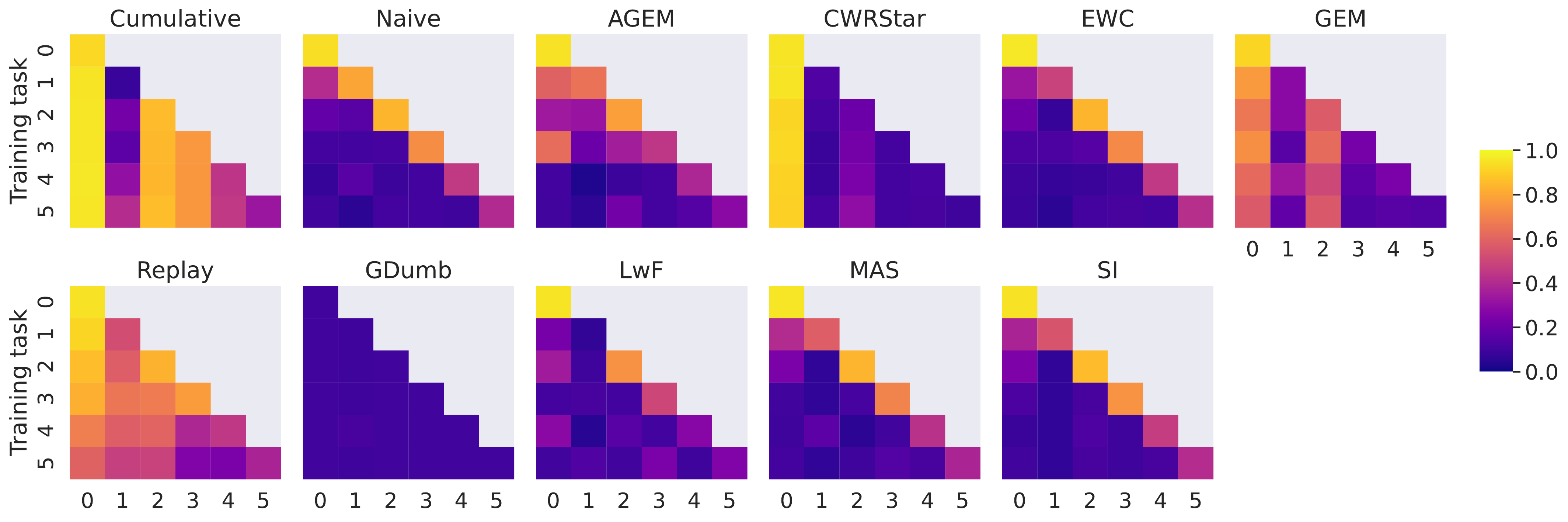}
        \caption{Experimental results (EfficientNet -- M2I -- Task-incremental) in terms of disaggregated performance ($\mathsf{ACC}$) on single tasks after learning previous tasks.}
        \label{fig:heatmaps_efficientnet_mnist_img_task}
    \end{figure}

\begin{figure}[h]
        \centering
        \includegraphics[width=\textwidth]{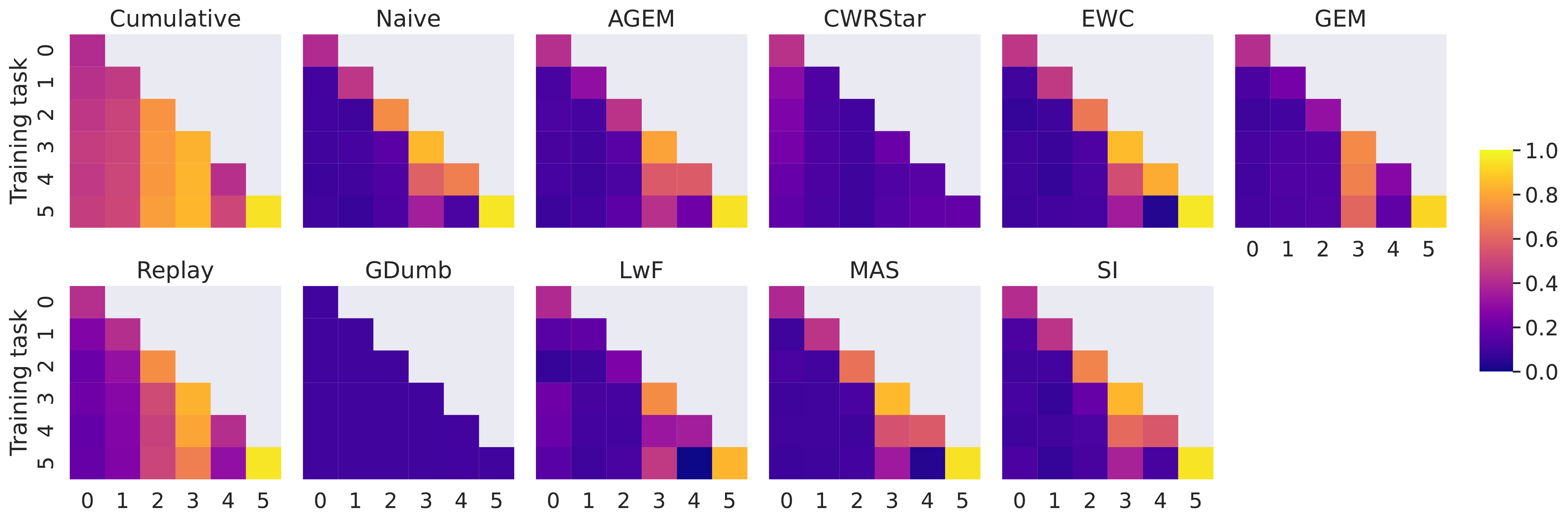}
        \caption{Experimental results (EfficientNet -- I2M -- Task-incremental) in terms of disaggregated performance ($\mathsf{ACC}$) on single tasks after learning previous tasks.}
        \label{fig:heatmaps_efficientnet_img_mnist_task}
    \end{figure}

 Shifting our focus to the heatmaps in Figure \ref{fig:heatmaps_efficientnet_mnist_img_class} -- \ref{fig:heatmaps_efficientnet_img_mnist_task} we are able to analyze in detail the forgetting of the different strategies throughout the experimental scenario. 

Figure \ref{fig:heatmaps_efficientnet_mnist_img_class} shows a vast amount of forgetting across all strategies. Some exceptions can be sparsely observed. For instance, MAS preserves its performance on task 0 after learning task 1, before dropping to values that are close to zero for previously encountered tasks. CWRStar preserves a remarkably high performance on the first task, but a very limited ability to incorporate new tasks. This result is in contrast with what was observed with VGG9, where the performance on the first task was preserved but decaying as new tasks are presented. This phenomenon may depend on the number of layers involved in the model backbone, since EfficientNet is a much larger model, and the weight adaptation strategy  used in CWRStar exclusively involves the last layer. As a result, a larger model such as EfficientNet will be more prone to knowledge retention than adaptation.
Moving to I2M in the class-incremental setting (see Figure \ref{fig:heatmaps_efficientnet_img_mnist_class}), a noteworthy result is GEM preserving knowledge of task 3 throughout the entire scenario, while not being able to preserve its performance on the other tasks. In this setting, CWRStar is fundamentally unable to learn any of the tasks presented in the scenario.

Interestingly, task similarity between two tasks, manifested by positive backward transfer, allows for improvement on previously learned tasks in some instances. In Figure \ref{fig:heatmaps_efficientnet_mnist_img_task}, for instance, learning task 4 is, in some cases, beneficial for the model's performance on task 1, as observed for MAS, GEM, and Naive. 

In the task-incremental setting (see Figure \ref{fig:heatmaps_efficientnet_img_mnist_task}), GEM presents a similar behavior to that observed in class incremental on task 2, but also preserves knowledge of task 0 throughout the entire scenario, whereas the performance on other tasks is fundamentally sub-optimal. In this setting, however, CWRStar behaves as in the M2I class incremental setting, i.e., the performance on task 0 is preserved throughout the entire scenario.


\subsubsection{Summary: RQ1}
In summary, results observed across the two learning settings (class-incremental, task-incremental) in the two presentation orders (M2I, I2M) show unsatisfactory performance for all learning strategies and that catastrophic forgetting is a real burden for many of the covered methods.

Considering that the results observed are inferior when compared to what is commonly reported in continual learning research, we can argue that our benchmark provides more challenging conditions for the CL strategies.
It is noteworthy that forgetting in CL strategies is also observed in perceptually similar tasks (e.g. MNIST, Omniglot, SVHN), as evident in our heatmaps. This behavior is indicative of the objective lack of robustness presented by CL strategies as they are exposed to tasks from different datasets.
The five desiderata described in Section \ref{sec:benchmark} and adopted to design our benchmarks set up a higher standard for the evaluation of CL strategies, and will hopefully stimulate the design and implementation of new, more robust strategies. 


\subsection{Discussion: RQ2}
In this subsection, we focus on the assessment of the ability of CL strategies to leverage curriculum task ordering to maximize their performance when exposed to our benchmarks devised in \ref{sec:benchmark}.






To answer this question, we start by analyzing results in Tables \ref{tab:results_vgg9_mnist_img} -- \ref{tab:efficient_net_not_pretrained_I2M}, which show metric values for the two scenarios: M2I (direct curriculum learning) and I2M (inverse curriculum learning). Almost all methods present a better performance in the curriculum learning setting (M2I) when compared with the inverse curriculum setting (I2M). 
Comparing values in Tables \ref{tab:results_vgg9_mnist_img} and \ref{tab:efficient_net_not_pretrained_M2I}, significant examples for VGG9 include Replay (from $0.550$ to $0.755$ in class-incremental and from $0.571$ to $0.730$ in task-incremental) and GEM (from $0.297$ to $0.572$ in class-incremental and from $0.269$ to $0.613$ in task-incremental).   Other CL strategies present a smaller margin of improvement. For instance, LwF (from $0.216$ to $0.222$ in class-incremental, and from 0.262 to $0.349$ in task-incremental) and EWC (from $0.190$ to $0.220$ in class-incremental, and from $0.340$ to $0.395$ in task-incremental).
Shifting the focus on results with EfficientNet  (comparing Tables \ref{tab:efficient_net_not_pretrained_M2I} and \ref{tab:efficient_net_not_pretrained_I2M}), examples include Replay (from $0.417$ to $0.655$ in class-incremental, and from $0.438$ to $0.605$ in task-incremental), and Cumulative (from $0.606$ to $0.834$ in class-incremental, and from $0.596$ to $0.656$ in task-incremental). Other CL strategies present a more limited but still significant improvement. For instance, CWRStar (from $0.053$ to $0.326$ in class-incremental, and from $0.171$ to $0.367$ in task-incremental), and GEM (from $0.218$ to $0.312$ in class-incremental, and from $0.276$ to $0.420$ in task-incremental). Counterexamples, where the model's performance is higher in the inverse curriculum setting (I2M), include LwF (from $0.203$ to $0.165$ in class-incremental and from $0.241$ to $0.240$ in task-incremental). This result shows that different strategies behave differently when presented with a different task ordering. 

Another interesting point pertaining to our research question is the opportunity to identify whether learning new tasks favors performance on previously learned tasks, emphasized in our heatmaps. This phenomenon may happen if the model is able to capture similarities between tasks that can be fruitfully leveraged for inference. To show some examples, we focus on task-incremental experiments. 
In the M2I scenario with VGG9, Figure \ref{fig:heatmaps_vgg9_mnist_img_task} shows that different strategies (AGEM, MAS, SI) are able to improve performance on task 1 (Omniglot) after learning task 5 (TinyImageNet). 
We also observe that multiple strategies (AGEM, EWC, SI, MAS) can leverage the skills learned in task 3 (SVHN) to improve performance on task 1 (MNIST). 
This result is intuitive since learning the complexity of street numbers in images acquired with a camera, strongly benefits the predictive capabilities on an easier dataset from the similar domain, i.e., MNIST. 
Similar behavior can be observed in Figure \ref{fig:heatmaps_efficientnet_mnist_img_task}, where AGEM improves the performance on task 2 (Fashion MNIST) after learning task 5 (TinyImagenet). 
In the I2M scenario with VGG9, Figure \ref{fig:heatmaps_vgg9_img_mnist_task} and Figure \ref{fig:heatmaps_efficientnet_img_mnist_task} show that almost all strategies improve the performance on task 3 (Fashion MNIST) and task 4 (Omniglot) after learning task 5 (MNIST).
This result shows that the knowledge learned from MNIST can boost the performance on more complex tasks learned before. 
Overall, results show that task ordering and task similarity can be leveraged to improve performance on a previously learned task, although the currently adopted CL strategies are sparsely able to yield this capability. This consideration paves the way for the design of new strategies that further leverage curriculum task ordering to boost forward and backward transfer.

An additional consideration pertains to the connection between curriculum learning and the appropriateness of CL metrics in this context. 
For instance, we note that MNIST is the simplest task and it is presented as the first task in the curriculum learning setting, it will be considered multiple times in the evaluation protocol, i.e., each time a new task is presented, which may boost the final average result presented by the accuracy metric. In turn, it is more likely for the curriculum learning setting to achieve higher average performance. This behavior poses issues in the interpretability of metric values, which are still unaddressed by currently available metrics. 
A clearer perspective is provided by the  heatmaps in Figures \ref{fig:heatmaps_vgg9_mnist_img_class}, \ref{fig:heatmaps_vgg9_img_mnist_class}, \ref{fig:heatmaps_vgg9_mnist_img_task}, \ref{fig:heatmaps_vgg9_img_mnist_task}.

\subsubsection{Summary: RQ2}

Overall, results observed across two learning settings (class-incremental, task-incremental) in the two presentation orders (M2I, I2M) show that current methods are not able to fully leverage curriculum learning. 
One reason may be the fact that most of the CL strategies are heavily impacted by forgetting since they are challenged by the complexities involved in our proposed benchmarks. 
Comparing the performance and behavior of the CL strategies between M2I and I2M scenarios, we can also observe that different task orderings significantly impact the final outcomes.
On the other hand, methods appear to partially benefit from task similarity in some specific cases, as highlighted in our analysis of results. This outcome leads us to the consideration that task similarity could be further exploited by CL strategies to yield models that simultaneously use the knowledge acquired from different tasks to perform better in every single task. 



\section{Conclusions}
In this work, we focused on the problem of benchmarking CL methods, which is often conducted in  heterogeneous ways, and with significant simplifications for the learning setting.
Specifically, we proposed two novel benchmarks that involve multiple heterogeneous tasks with varying qualities and complexities.
Our benchmarks involved six image datasets in increasing (M2I) and decreasing (I2M) difficulty order, following the curriculum learning paradigm. 
The heterogeneity across datasets allowed us to inject realistic complexities into the learning scenario, resulting in challenging conditions for CL strategies.
Particular emphasis was put on the rigorous and reproducible evaluation of model generalization capabilities and forgetting.
Our extensive experimental evaluation showed that popular CL strategies, which are known to be robust on commonly adopted scenarios, fail to achieve satisfactory performance with our benchmarks. Moreover, CL strategies are affected by forgetting and are not able to effectively leverage curriculum task ordering to improve their performance and robustness, missing on the opportunity of simultaneously using knowledge from different tasks to perform better in every single task.
Our results represent a starting point to assess the impact of curriculum learning on CL strategies. 
Future work includes the design of new CL strategies that are able to deal with the complexities devised in our benchmarks. Moreover, from an evaluation perspective, new metrics could be investigated to fully capture the spectrum of model behavior with different task orderings. 
Finally, an interesting line of research pertains to the analysis of the behavior of non-conventional CL strategies, which are not yet incorporated in known frameworks due to their emerging nature.

\newpage 
\section*{Declarations}


\begin{itemize}
\item \textbf{Funding} -- The paper was supported by the Polish Ministry of Science and Higher Education allocated to the AGH UST. \\

\item \textbf{Conflict of interest/Competing interests} -- All authors certify that they have no affiliations with or involvement in any organization or entity with any financial interest in the subject matter or materials discussed in this manuscript. \\

\item \textbf{Ethics approval} -- Not applicable. \\

\item \textbf{Consent to participate} -- This study does not involve human subjects or any sensitive data. \\

\item \textbf{Consent for publication} -- This study does not involve human subjects or any sensitive data.  \\

\item \textbf{Availability of data and materials} -- The data and materials to reproduce the experiments are available at the following repository URL: \url{https://github.com/lifelonglab/M2I_I2M_benchmark} \\

\item \textbf{Code availability} -- The code of the proposed benchmarks is available at the following repository URL: \url{https://github.com/lifelonglab/M2I_I2M_benchmark} \\

\item \textbf{Authors' contributions} -- Kamil Faber: \textit{Data Curation, Investigation, Software, Visualization, Writing} -- Dominik Zurek: \textit{Data Curation, Investigation, Resources, Software, Writing (Review \& Editing)} -- Marcin Pietron, Nathalie Japkowicz: \textit{Resources, Validation, Writing (Review \& Editing)} -- Antonio Vergari, Roberto Corizzo: \textit{Conceptualization, Supervision, Methodology, Project Administration, Writing} \\
\end{itemize}

\noindent

\bibliographystyle{sn-mathphys}
\bibliography{referomnia}

\end{document}